\documentclass[conference]{IEEEtran}
\usepackage{cite}
\usepackage{amsmath,amssymb,amsfonts}
\usepackage{graphicx}
\usepackage{textcomp}
\usepackage{xcolor}
\usepackage{hyperref}
\usepackage{url}
\usepackage{titlesec}
\usepackage{tikz}
\usepackage{setspace} 
\usepackage{enumitem}
\usepackage{caption}
\usepackage{subcaption}
\usepackage[normalem]{ulem}
\usepackage{pifont}
\usepackage{multirow}
\usepackage{algorithm}
\usepackage{algorithmicx}
\usepackage{algpseudocode}
\usepackage{mathtools, nccmath}
\usepackage{xspace}
\usepackage{adjustbox}
\usepackage{booktabs}

\usepackage{caption}

\newcommand{\xmark}{\ding{55}}%
\newcommand{\cmark}{\ding{51}}
\DeclarePairedDelimiter{\nint}\lfloor\rceil

\def\BibTeX{{\rm B\kern-.05em{\sc i\kern-.025em b}\kern-.08em
    T\kern-.1667em\lower.7ex\hbox{E}\kern-.125emX}}

\newcommand{\ourmethod}{ALISA\xspace}

\newcommand{\minisection}[1]{\vspace{0.05in}\noindent {\bf #1}}

\algrenewcommand\algorithmicrequire{\textbf{Input:}}
\algrenewcommand\algorithmicensure{\textbf{Output:}}

\title{ALISA: \underline{A}ccelerating \underline{L}arge Language Model \underline{I}nference via \underline{S}parsity-\underline{A}ware KV Caching}

\author{\IEEEauthorblockN{Youpeng Zhao,
Di Wu, Jun Wang}
\IEEEauthorblockA{University of Central Florida \\
Email: \{youpeng.zhao, di.wu, jun.wang\}@ucf.edu
}}

\begin{document}
\maketitle
\thispagestyle{plain}
\pagestyle{plain}


\begin{abstract}
The Transformer architecture has significantly advanced natural language processing (NLP) and has been foundational in developing large language models (LLMs) such as LLaMA and OPT, which have come to dominate a broad range of NLP tasks. 
Despite their superior accuracy, LLMs present unique challenges in practical inference, concerning the compute and memory-intensive nature. 
Thanks to the autoregressive characteristic of LLM inference, KV caching for the attention layers in Transformers can effectively accelerate LLM inference by substituting quadratic-complexity computation with linear-complexity memory accesses.
Yet, this approach requires increasing memory as demand grows for processing longer sequences.
The overhead leads to reduced throughput due to I/O bottlenecks and even out-of-memory errors, particularly on resource-constrained systems like a single commodity GPU.

In this paper, we propose \ourmethod, a novel algorithm-system co-design solution to address the challenges imposed by KV caching. 
On the algorithm level, \ourmethod prioritizes tokens that are most important in generating a new token via a Sparse Window Attention (SWA) algorithm.
SWA introduces high sparsity in attention layers and reduces the memory footprint of KV caching at negligible accuracy loss.
On the system level, \ourmethod employs three-phase token-level dynamical scheduling and optimizes the trade-off between caching and recomputation, thus maximizing the overall performance in resource-constrained systems.
In a single GPU-CPU system, we demonstrate that under varying workloads, \ourmethod improves the throughput of baseline systems such as FlexGen and vLLM by up to $3\times$ and $1.9\times$, respectively.

\end{abstract}
\section{Introduction}
Large Language Models (LLMs) stand as a revolutionary breakthrough in the modern era of artificial intelligence (AI). 
Distinct from previous small language models with only millions of parameters, LLMs often have hundreds of billions or even trillions of parameters. 
They have exhibited exceptional abilities in solving complex tasks, such as semantic reasoning and creative writing through text generation.
GPT-2 XL, one of the earliest LLMs with 1.5 billion parameters, pioneered in showcasing these capabilities~\cite{gpt2}.
Its successor, GPT-3, showcases even more powerful abilities with 175 billion parameters~\cite{gpt3}.
To date, the most noteworthy application of LLMs is ChatGPT from OpenAI~\cite{chatgpt}, a tool that allows users to interact with an AI agent in a conversational way to solve tasks ranging from language translation to software engineering, and beyond.

\begin{figure}[!t]
\begin{center}
  \includegraphics[width=.95\linewidth]{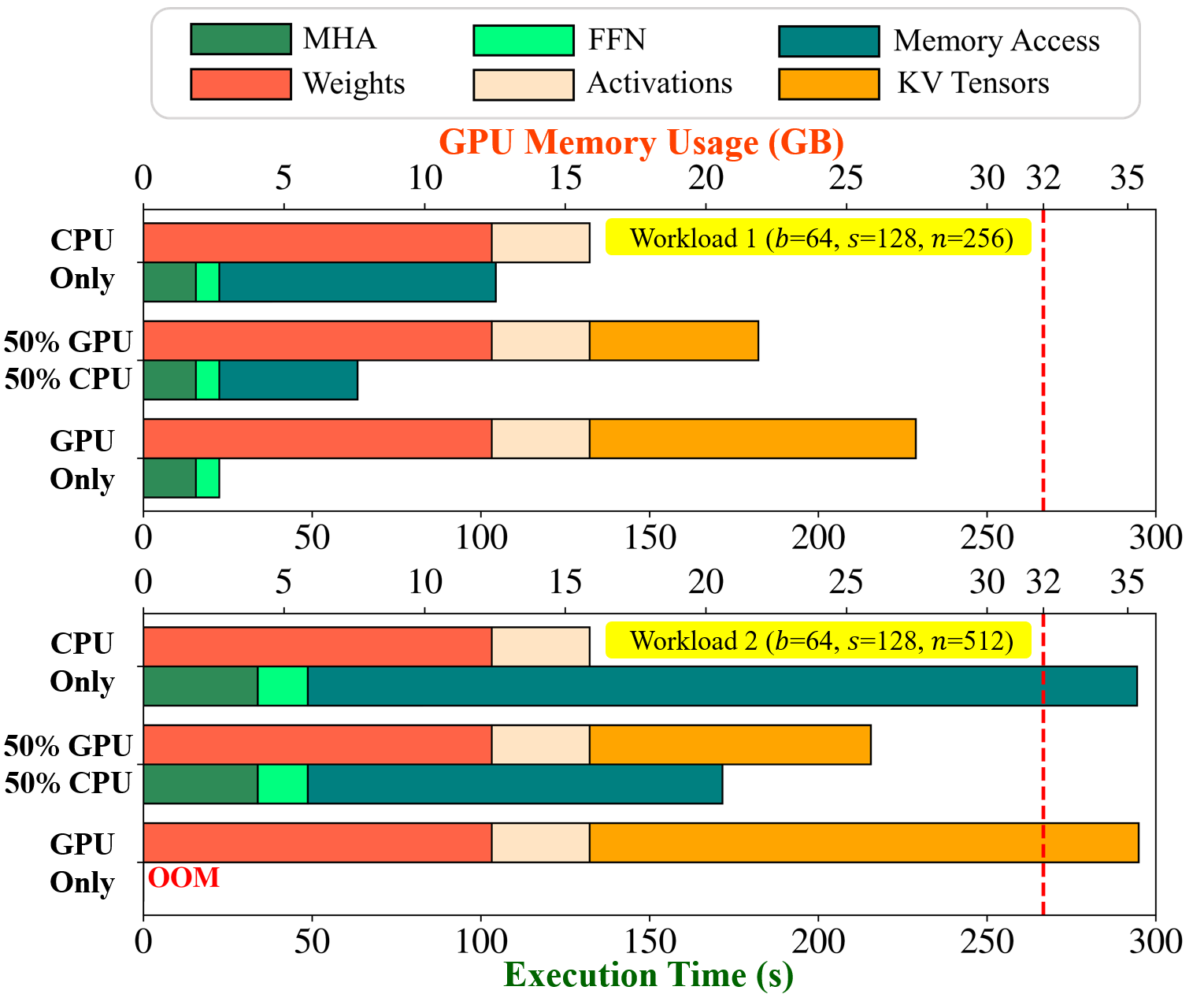}
\end{center}
\vspace{-3mm}
  \caption{Breakdown of execution time and memory usage for OPT-6.7B inference on one NVIDIA Tesla V100 GPU with 32~GB memory under different workloads. 
  Weights, activations, and KV tensors (intermediate key and value states) denote the required GPU memory.
  MHA, FFN, and memory access denote the time for computing multi-headed attention, and feed-forward network (both including the follow-up Addition and LayerNorm operations) and KV caching (moving KV tensors between CPU and GPU if any). 
  50\% means the ratio of the KV tensors allocated to CPU/GPU memory.
  OOM denotes out-of-memory error, and the red-dot line denotes the GPU memory capacity.
  The $b$, $s$, and $n$ for workloads refer to the batch size, and input and output sequence length. Results are reported using FlexGen~\cite{flexgen}.
  }
\label{fig:intro}
\end{figure}

\begin{figure*}[!ht]
\begin{center}
  \includegraphics[width=\linewidth]{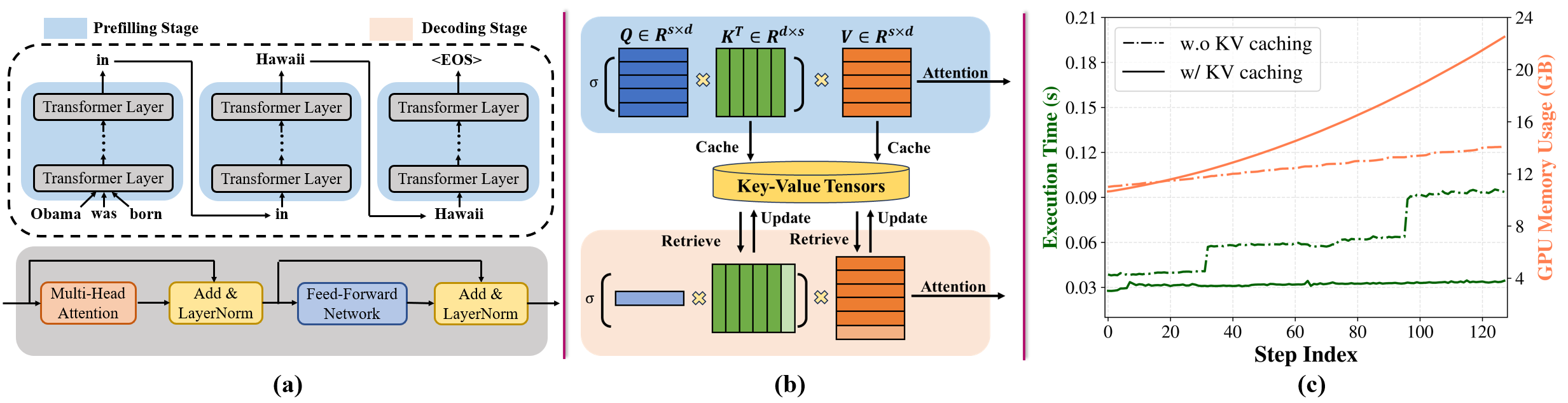}
\end{center}
\vspace{-3mm}
  \caption{(a) Top: an example of autoregressive LLM inference. EOS refers to end-of-sentence.
  Bottom: operation blocks in transformer layers.
  (b) KV caching: 
  $Q$, $K$, $V$ denotes the query, key, and value tensors.
  At the prefilling stage, all input tokens are processed simultaneously, and the generated intermediate KV tensors are stored, marked by dark colors. 
  $s$ and $d$ represent the input sequence length and the hidden dimension size of KV tensors. 
  At the decoding stage, the stored KV tensors in the dark colors are retrieved. 
  The input $Q$, $K$, $V$ tensors are marked by the light colors. 
  The input $Q$ tensor is multiplied with a concatenation of input $K$ and stored $K$ tensors, followed by a softmax of the entire attention weights.
  The attention weight are further multiplied with a concatenation of input $V$ and stored $V$ tensors to generate new results.
  Afterward, the input $K$ and $V$ tensors are stored.
  This process is repeated per token. 
  (c) Execution time and GPU memory usage for OPT-6.7B inference with and without KV caching. 
  The x-axis step index means the output sequence length.
  Results are reported using HuggingFace Accelerate~\cite{huggingface}.
  }
\label{fig:kv}
\end{figure*}

LLMs usually consist of stacked transformer decoder layers, in which the critical component is self-attention (attention for short in this work)~\cite{attention}.
The attention modules empower LLMs to capture contextual information by attending to different positions within the sequences, which however introduces quadratic computation complexity with the sequence length.
Such complexity severely bottlenecks the performance and scalability of LLMs and is becoming more pronounced upon the pursuit for longer sequences in existing systems~\cite{longformer, sparseformer, esti, longnet}.
One viable solution to this problem during LLM inference is KV caching~\cite{fairseq}.
This idea originates from the fact that LLM inference is \textit{autoregressive}, where LLMs generate new tokens sequentially based on all prior tokens (more details in Figure~\ref{fig:kv}).
This characteristic opens up the opportunity of reusing intermediate states, specifically, the key ($K$) and value ($V$) tensors, whose sizes increase linearly with the sequence length through caching in attention layers.
With KV caching, the quadratic-complexity computation is reduced to linear-complexity computation and memory accesses, therefore substantially speeding up LLM inference.

\minisection{Challenge.}
Despite KV caching significantly reducing the inference time, \textit{LLM inference with KV caching is predominantly bottlenecked by memory~\cite{esti}, especially in resource-constrained systems}, like a single commodity GPU.
First, the most expensive operations in LLMs are matrix multiplication and softmax, which are notoriously memory-bound.
Second, the weights and activations in LLMs have already raised an alert on the memory capacity.
Third, intermediate KV tensors further exacerbate the requirement on memory capacity, which is determined by the sequence length and model dimension.
For a given LLM, as the batch size and sequence length increase, the allocated memory for KV caching continues to grow linearly, and at some point, exceeds the available memory capacity.
Ultimately, pursuing longer sequences in larger LLMs ends up with an out-of-memory error and halts the execution, as given by the ``GPU only'' case on workload 2 in Figure~\ref{fig:intro}.
To circumvent the out-of-memory error in a single GPU setting, researchers have developed solutions to offload KV tensors to CPU memory or even secondary storage to free up GPU resources in real-world scenarios~\cite{flexgen}. 
However, frequent offloading and reloading of KV tensors incur significant data transfer overhead, which becomes the new bottleneck towards high throughput and low latency in resource-constrained systems, as seen in Figure~\ref{fig:intro}.
To this end, we ask: \textit{how to innovate KV caching for LLMs in resource-constrained systems, to facilitate better scalability and meet the need of longer sequences and larger model sizes.}

\minisection{Proposal.}
In this paper, we propose \ourmethod, an algorithm-system co-design solution to \underline{a}ccelerate \underline{L}LM \underline{i}nference via \underline{s}parsity-\underline{a}ware KV caching for single GPU-CPU systems. 
Our key observation is that \textit{during the autoregressive inference process, the attention weight matrix is highly sparse, and larger LLMs exhibit higher attention weight sparsity.}
This observation validates the intuition that not all tokens are created equal and only a small number of important tokens contribute to generating a new token.
Once these important tokens are identified, we can selectively access the KV tensors corresponding to these important tokens, and skip unimportant ones.
To identify which tokens are important, we formulate a Sparse Window Attention (SWA) algorithm, in which both the globally dynamic and locally static sparse patterns are created.
A mixture of these sparse patterns can significantly reduce the memory footprint while maintaining model accuracy due to the ability to better capture important tokens in a sequence.

However, as the size of LLMs keeps growing, the above algorithmic optimization is insufficient to guarantee satisfactory performance, i.e., throughput in this work, for resource-constrained systems.
We argue that \textit{accelerating LLMs is not only a computation problem but more of a memory problem, in the presence of a gigantic memory footprint.}
Three bottlenecks are responsible.
Firstly, the size of sparse KV tensors will ultimately exceed the memory capacity with longer sequences, and the long-latency GPU-CPU memory accesses in dense LLMs recur.
Secondly, the sparse nature of KV tensors induces unpredictable memory access, which is exacerbated by longer sequences.
To address these two challenges, we propose to dynamically schedule the KV tensors at the token level and balance between caching and recomputation for best performance gain.
We highlight this token-level scheduling in Table~\ref{tab:1}.
Thirdly, high-precision (FP16 in this work) KV tensors still exhibit a large memory footprint, thus high memory access latency.
We can compress KV tensors to lower precision (INT8) via quantization and further reduce the overall memory overhead, without sacrificing the accuracy.

\begin{table}[!t]
\begin{center}
\caption{Comparison of prior works and our \ourmethod. Block means a fixed group of tokens. Head means a single attention module.}
\resizebox{\columnwidth}{!}{\begin{tabular}{c||c|c|c}
\toprule
\textbf{Design} & \textbf{vLLM~\cite{vLLM}} & \textbf{FlexGen~\cite{flexgen}} & \textbf{\ourmethod (Ours)} \\ \midrule\midrule
Sparse Attn. & \xmark & \xmark & \cmark \\ \midrule
\begin{tabular}[c]{@{}c@{}}Caching \\ Granularity\end{tabular} & \begin{tabular}[c]{@{}c@{}}Block-level \\ (Static)\end{tabular} & \begin{tabular}[c]{@{}c@{}}Head-level \\ (Static)\end{tabular} & \begin{tabular}[c]{@{}c@{}}Token-level \\ (Dynamic)\end{tabular} \\ \midrule
Recomputation & \cmark & \xmark & \cmark \\ \midrule
\begin{tabular}[c]{@{}c@{}}Scenario\end{tabular} & \begin{tabular}[c]{@{}c@{}}Online \\ (Multi-GPU)\end{tabular} & \begin{tabular}[c]{@{}c@{}}Offline \\ (Single-GPU)\end{tabular} & \begin{tabular}[c]{@{}c@{}}Offline \\ (Single-GPU)\end{tabular} \\ \midrule
\begin{tabular}[c]{@{}c@{}}Co-Design\end{tabular} & \xmark & \xmark & \cmark \\ \bottomrule
\end{tabular}
\label{tab:1}}
\end{center}
\end{table}

In summary, this paper makes the following contributions: 
\begin{itemize}
        \item We identify the challenges in KV caching for LLM inference and propose an algorithm-system co-design solution, \ourmethod, for efficient LLM inference in resource-constrained systems.
        \item On the algorithm level, we propose sparse window attention (SWA) that creates a mixture of globally dynamic and locally static sparse patterns in KV tensors to reduce the memory footprint while maintaining high accuracy.
        \item On the system level, we design a three-phase scheduler to dynamically allocate KV tensors between GPU and CPU memory to reduce data transfer at the token level.
        \item We evaluate \ourmethod over a wide range of LLM models, tasks, and workloads. 
        Experiments demonstrate that \ourmethod can significantly reduce the memory footprint of KV tensors and increase the throughput over previous baselines, with negligible accuracy drop.
\end{itemize}

The remainder of this paper is organized as follows. Section~\ref{sec:background} recaps LLM-related concepts and works.
Then, Section~\ref{sec:challenge_opportunity} articulates our perspectives on challenges and corresponding opportunities in accelerating LLMs.
Next, Section~\ref{sec:algorithm_design} and Section~\ref{sec:system_design} elaborate the algorithm and system design in \ourmethod, with evaluation followed in Section~\ref{sec:evaluation}.
Finally, Section~\ref{sec:conclusion} concludes this work.

\section{Background}\label{sec:background}
In this section, we first present some preliminary knowledge of LLMs, including autoregressive inference, the Transformer layer, and KV caching. 
Afterwards, we discuss related works.
\subsection{Large Language Models}
\minisection{Autoregressive Inference.} Transformer-based language models function by processing a sequence of input words and generating new, related words as output.
Compared with previous small language models in the pre-LLM era, the most distinctive characteristic of LLMs is that LLM inference is \textit{autoregressive}, i.e., output tokens solely depend on past tokens.
Figure~\ref{fig:kv}~(a) gives an example of such an autoregressive behavior in LLM inference on the top.
The inference process can be divided into two parts, including the prefilling and decoding stages.
During the prefilling stage, LLMs process all the input tokens in a single pass.
Then, during the decoding stage, a previously generated output token is fed back into the model as an input and generates the next output token. 
Therefore, the decoding stage unfolds iteratively, processing one token at a time.
When the sequence length reaches a maximum threshold (specified by users or service providers) or an ``$\langle$EOS$\rangle$'' token is emitted, the decoding process stops.
Inside the LLMs, the input words are first tokenized to continuous vectors using an embedding layer (not shown for simplicity) and then go through stacked transformer layers.
All transformer layers are identical and include a multi-head attention (MHA) layer and a feed-forward network (FFN) layer, as shown at the bottom in Figure~\ref{fig:kv}~(a).
There exist addition and layer normalization layers after MHA and FFN layers.
We consider them as part of the MHA and FFN layers in this work during evaluation.
Finally, the outputs of transformer layers will go through a linear projection and a softmax layer to generate the corresponding token for the next word (not shown).

\minisection{Transformer Layer.} At the core of transformer layers lies the attention module~\cite{attention}.
The relevant operations are given in Equation~\ref{eq:1} and~\ref{eq:2}.
There are three intermediate tensors involved, namely, query $Q$, key $K$, and value $V$. 
The attention weights $\textit{AW}(Q, K)$ are calculated by first computing the dot product between $Q$ and $K$, then scaling the product by the square root of hidden dimension $d$, and finally going through a softmax operation ($\sigma(\cdot)$).
The attention scores $\textit{Attn}(Q, K, V)$ are calculated by multiplying the attention weights $\textit{AW}(Q, K)$ to $V$.
The MHA output is obtained by simply concatenating the outputs of all attention heads along the head dimension, with each head being an attention module.
\begin{align}
    \textit{AW}(Q, K) &= \sigma(\frac{QK^{T}}{\sqrt{d}}) \label{eq:1} \\
    \textit{Attn}(Q, K, V) &= \textit{AW}(Q, K) \cdot V \label{eq:2}
\end{align}

\minisection{KV Caching.} According to Equation~\ref{eq:1}, the attention operation induces quadratic computation complexity with respect to the sequence length.
An example is given at the top of Figure~\ref{fig:kv}~(b). 
The sequence length quadratically increases the size of the attention weight matrix, therefore quadratically increasing execution time.
This overhead is exacerbated when pursuing longer sequences for larger models~\cite{longformer, sparseformer, esti, longnet}.
To mitigate such a quadratic overhead for LLM inference, KV Caching is proposed to store the intermediate tensors such as key ($K$) and value ($V$) tensors in attention layers for computation reuse in future decoding steps~\cite{fairseq}.
The bottom of Figure~\ref{fig:kv}~(b) showcases how KV caching works.
KV caching transforms the original matrix multiplication with quadratic complexity into vector-matrix multiplication and memory accesses with linear complexity, thus significantly improving the performance.
Figure~\ref{fig:kv}~(c) draws the execution time and memory usage for LLM inference with and without KV caching.
Without KV caching, the execution time increases rapidly, due to calculating the entire attention repeatedly.
With KV caching, only the attention weights and scores for the newly generated token are calculated as vector-matrix multiplication, and the execution time stays almost constant across different steps. 
However, such runtime reduction is at the expense of GPU memory usage, which increases gradually over time, due to the growing size of KV tensors.

\subsection{Related Work}
\minisection{Algorithmic Optimization for Attention.}
On the algorithm side, various optimizations have been proposed to address the quadratic complexity of attention modules.
In the pre-LLM era, algorithm optimizations largely focus on reducing the attention complexity through approximation methods.
For example, Linformer~\cite{linformer} and Reformer~\cite{reformer} approximate the original attention using low-rank matrices and locality-sensitive hashing, respectively, achieving almost linear complexity. 
However, these approximations are not able to offer competitive accuracy in LLMs.
Another line of algorithmic optimization is to create sparse patterns in attention modules~\cite{sparsebert,efficientatten,longformer,sparseformer}.
However, most sparsity-driven methods require additional training, which is not scalable for LLMs and cannot guarantee accuracy performance~\cite{sparsebert,efficientatten}.
In the LLM era, Longformer~\cite{longformer} constructs the sparse attention using a fixed-size sliding window on the most recent local tokens. 
SparseTransformer~\cite{sparseformer} generates sparse patterns with a fixed stride on all tokens. 
However, these sparse attention methods are not able to capture important tokens during the autoregressive LLM inference, resulting in accuracy collapse.

\minisection{Hardware Acceleration for Attention.}
For small language models, accelerators that utilize algorithm-hardware co-design have been proposed~\cite{spatten,vitality}. 
For example, SpAtten co-designs the algorithm and accelerator architecture to improve the sparsity in attention modules and reduce both the compute and memory overheads in matrix multiplication operations~\cite{spatten}; 
ViTALiTy approximates the dot-product softmax operation in attention modules using first-order Taylor expansion and linearizes the relevant cost~\cite{vitality}.
Though these accelerators are quite effective in the pre-LLM era, their merit fades away in LLM inference, due to their fundamental limitations. 
First, pre-LLM accelerators simply can not handle the large model size of LLMs.
For example, SpAtten balances its design choices among algorithm complexity, computation throughput, and memory capacity for the BERT~\cite{bert}, GPT-2 small and medium model~\cite{gpt2}.
However, the largest GPT-2 medium model has only 0.36~billion parameters, not even a fraction of that for LLMs, e.g., 175~billion parameters for GPT-3~\cite{gpt3}, which engages 652~GB for single-precision model weights.
Naively slabbing large memory onto the computing kernels does not offer Pareto efficiency.
Second, prior co-designed accelerators are not able to further scale up with longer sequences.
For example, SpAtten requires storing the entire attention weights to prune away unwanted tokens.
However, the size of the attention weight matrix increases quadratically with sequence length.
In the era of LLMs, given limited memory capacity, especially in resource-constrained systems, squeezing memory from KV tensors to attention weights will certainly degrade the efficacy of KV caching and slow down the inference.

\minisection{KV Caching Optimization.} In the LLM era, numerous specialized LLM systems have been developed. 
We compare some of these systems in Table~\ref{tab:1}.
For example, FlashAttention aims to reduce memory accesses between on-chip SRAM and off-chip HBM in GPUs for higher throughput with fine-grained tiling and partitioning at the kernel level~\cite{flashattention,flashattention2}.
However, it does not optimize the memory accesses between CPUs and GPUs.
vLLM proposes storing intermediate KV tensors at the block level, where each block contains a fixed number of tokens and is stored in non-contiguous paged memory to alleviate memory fragmentation for online LLM inference~\cite{vLLM}.
Identical to this work, FlexGen also targets resource-constrained systems~\cite{flexgen}.
FlexGen formulates a static offloading strategy for KV tensors throughout the LLM inference and manages them at the head level.
$H_{2}O$ designs a KV caching policy by retaining heavy hitters ($H_2$) tokens, which are determined by the global attention weight sum~\cite{h2o}, rather than the local attention weight sum in \ourmethod.
To summarize, three features differentiate \ourmethod from prior works.
First, \ourmethod co-designs both the algorithm and system to fully exploit sparse attention for higher throughput, while previous works focus solely on either algorithm improvement ($H_{2}O$) or system improvement (vLLM, FlexGen).
Second, \ourmethod performs KV caching at the granularity of one token, allowing flexible KV tensor allocation, which is critical upon sparsity-driven co-design.
Third, \ourmethod adopts an appropriate dynamic scheduler to perform both caching and recomputation, while previous works only employ static KV caching~\cite{flexgen,h2o}.

\section{Challenges and Opportunities}\label{sec:challenge_opportunity}
\subsection{Challenges}
Despite KV caching has significantly improved the end-to-end performance for LLMs by avoiding quadratic-complexity computation, it still introduces a linear-complexity memory footprint.
During LLM inference, we have to allocate GPU memory to store intermediate KV tensors. 
The corresponding memory footprint can vary from hundreds of megabytes to hundreds of gigabytes, depending on batch size, sequence length, and model configuration. 
As shown in Figure~\ref{fig:kv}~(c), the GPU memory usage with KV caching is about 60\% higher than that without KV caching with only 128 tokens in the sequence.
In half-precision data format, running OPT-13B with a sequence length of 512 at a batch size of 64 imposes more than 25~GB of memory for KV tensors, which is even larger than the model weight size (about 23~GB).
When the sequence length increases, this gap will be further widened.

In resource-constrained systems (e.g., a single GPU with limited memory), KV tensors ought to be offloaded to next-level memory hierarchies, such as CPU memory or even secondary storage, when the size of KV tensors exceeds the capacity of the GPU memory. 
However, offloading and reloading KV tensors incur significant data transfer overhead (e.g., I/O access between GPU and CPU memory on the PCIe bus), as shown in Figure~\ref{fig:intro}.
Storing 50\% KV tensors in CPU memory will increase the overall execution time of LLM inference by $3\times$;
and this slowdown reaches $5\times$ if storing all KV tensors in CPU memory.
Given this bottleneck in KV caching, we need to find \textit{a solution that orchestrates when, how, and what to offload and reload in resource-constrained systems, so that the overall execution time is minimized.}

\subsection{Opportunities} 
\begin{figure}[!t]
\begin{center}
  \includegraphics[width=\linewidth]{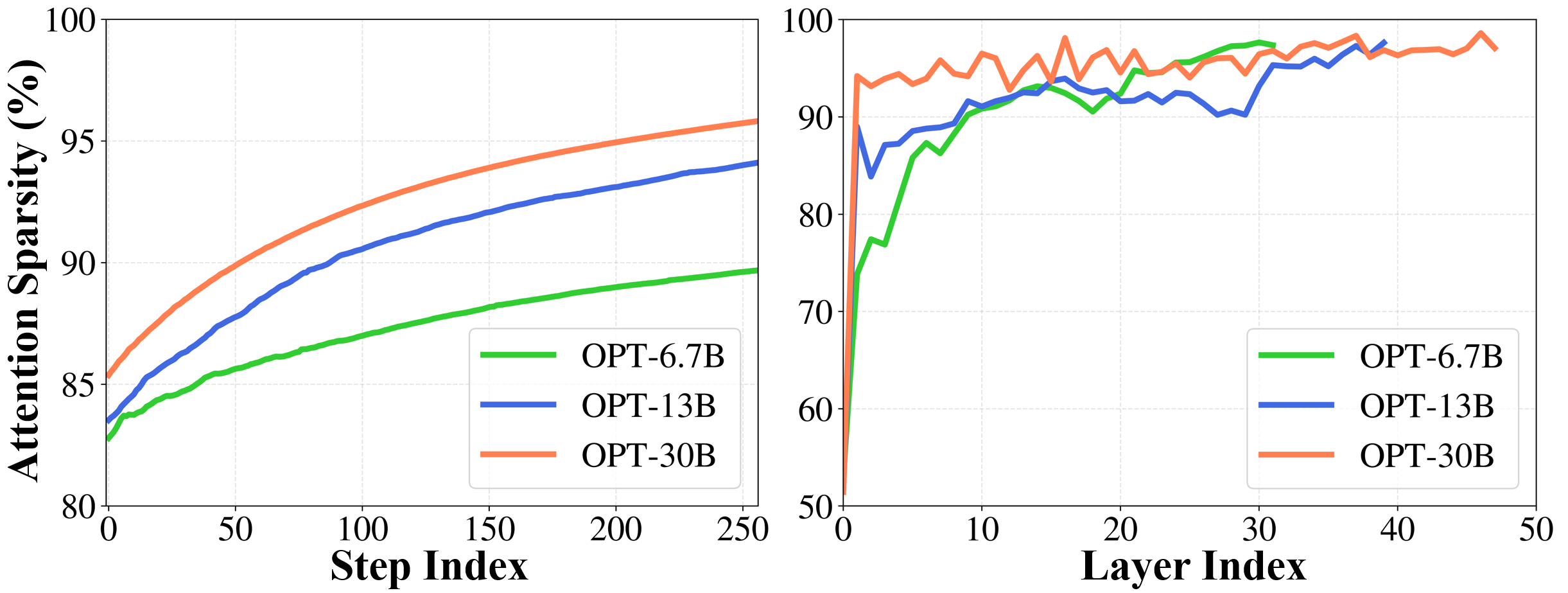}
\end{center}
\vspace{-2mm}
  \caption{Attention weight sparsity observed across different steps and layers during OPT model inference using the Wiki-Text-2 dataset~\cite{wiki}. We consider elements as zeros if they fall below 1\% of the row-wise maximum value.
  }
\label{fig:sps}
\end{figure}

Let's start with a simple example.
Given the question ``What is the capital of France?'', we humans only need to pay attention to `capital' and `France' to respond with the answer ``Paris.''
The intuition is that not all words (tokens) are created equal, and some are more important than others.
This intuition has been leveraged in accelerating transformers in the pre-LLM era.
Prior works for small language models empirically keep the tokens that lead to large attention weights, and prune away those with smaller weights~\cite{spatten}.
In this work, we take one more step to corroborate this intuition in LLMs by profiling the sparsity in attention weights, as shown in Figure~\ref{fig:sps}.
We have two key observations.
First, the attention weights in LLMs are highly sparse, e.g., the sparsity can vary between 80\% and 95\% across different inference steps, and reach close to 99\% in some layers. 
Second, larger LLMs exhibit higher sparsity, e.g., the density (i.e., $1 - \text{sparsity}$) of OPT-30B is about $3\times$ less than that of OPT-6.7B.
These observations translate to the fact that, from a computation perspective, very few elements in the attention weight matrix contribute to calculating the final attention score and generating new tokens in LLMs. 
This both motivates and validates our solution to create sparse KV tensors by skipping unimportant tokens in LLM inference.

\begin{figure*}[!t]
     \centering
     \begin{subfigure}[b]{0.245\linewidth}
         \centering
         \caption{Dense Attention}
         \includegraphics[width=\linewidth]{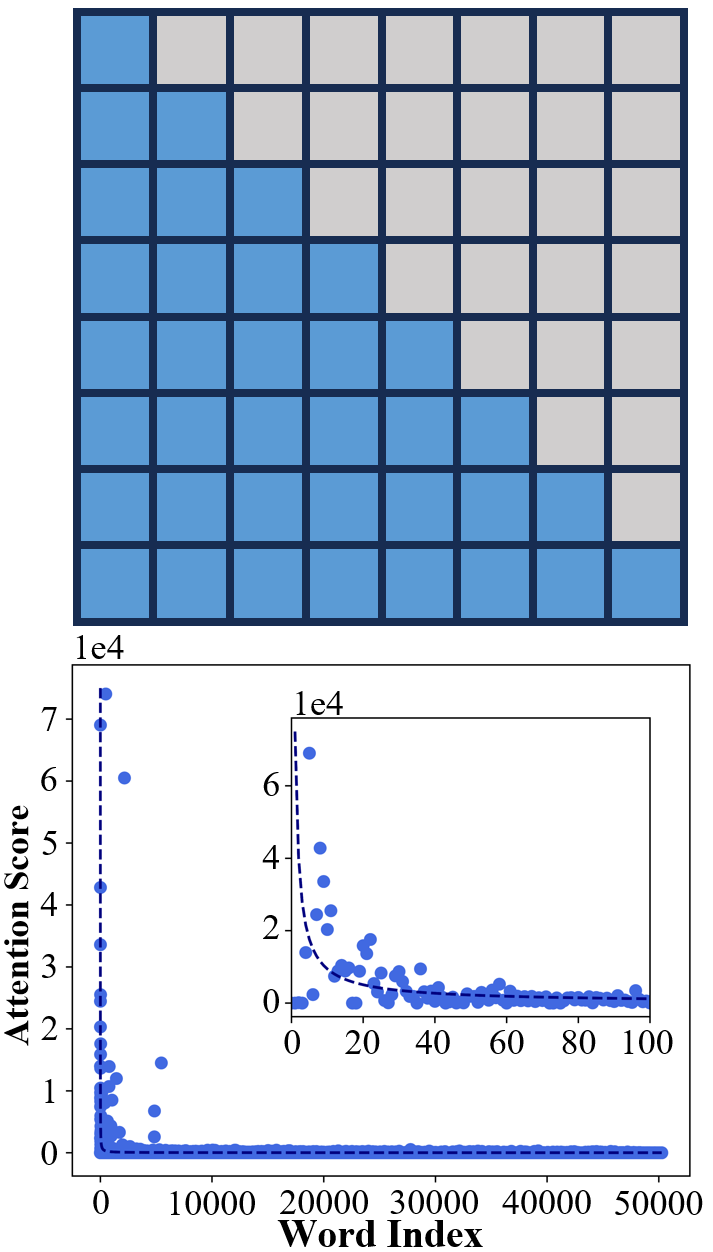}
         \label{fig:attn-1-dense}
     \end{subfigure}
     \hfill
     \begin{subfigure}[b]{0.245\linewidth}
         \centering
         \caption{Local Attention~\cite{longformer}}
         \includegraphics[width=\linewidth]{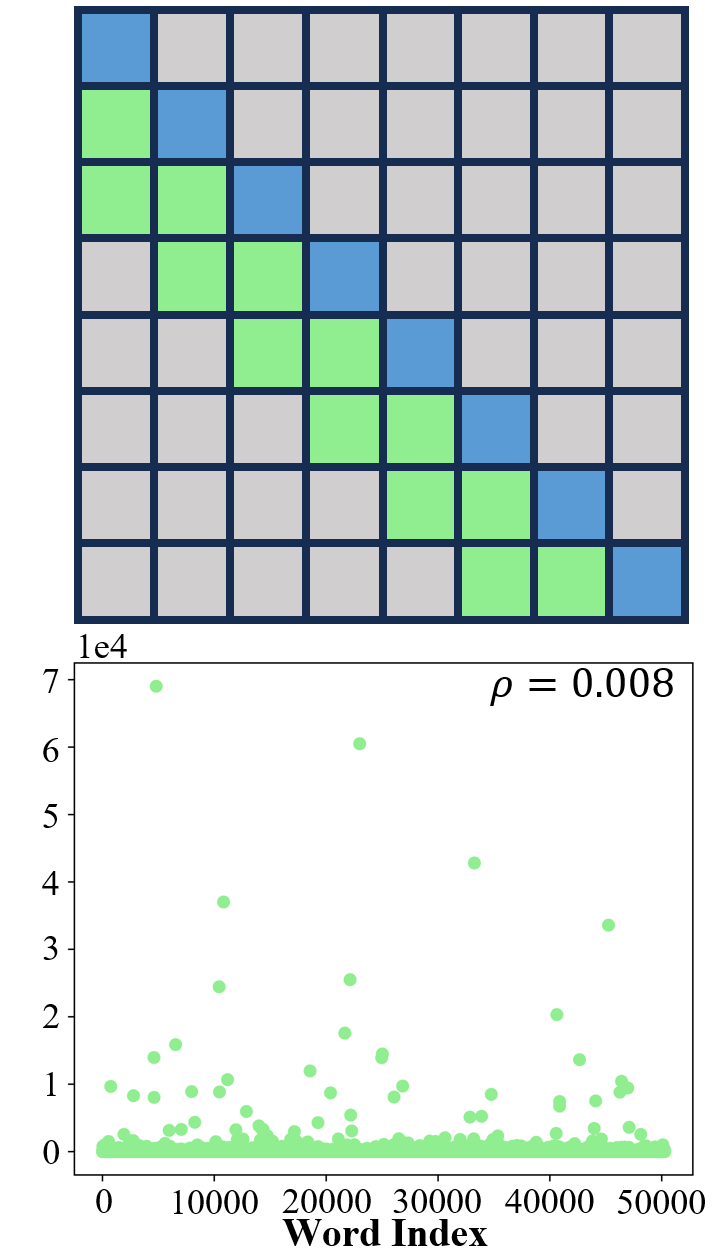}
         \label{fig:attn-1-local}
     \end{subfigure}
     \hfill
     \begin{subfigure}[b]{0.245\linewidth}
         \centering
         \caption{Strided Attention~\cite{sparseformer}}
         \includegraphics[width=\linewidth]{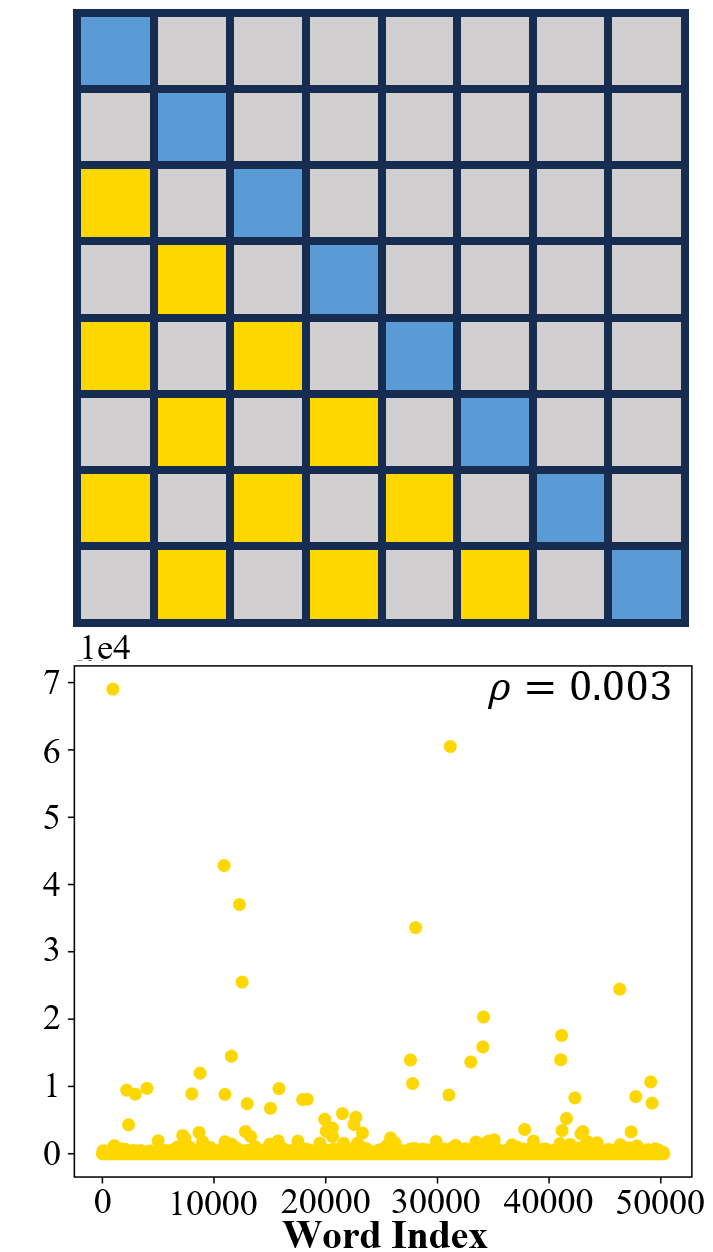}
         \label{fig:attn-1-stride}
     \end{subfigure}
     \hfill
     \begin{subfigure}[b]{0.245\linewidth}
         \centering
         \caption{\textbf{SWA (Ours)}}
         \includegraphics[width=\linewidth]{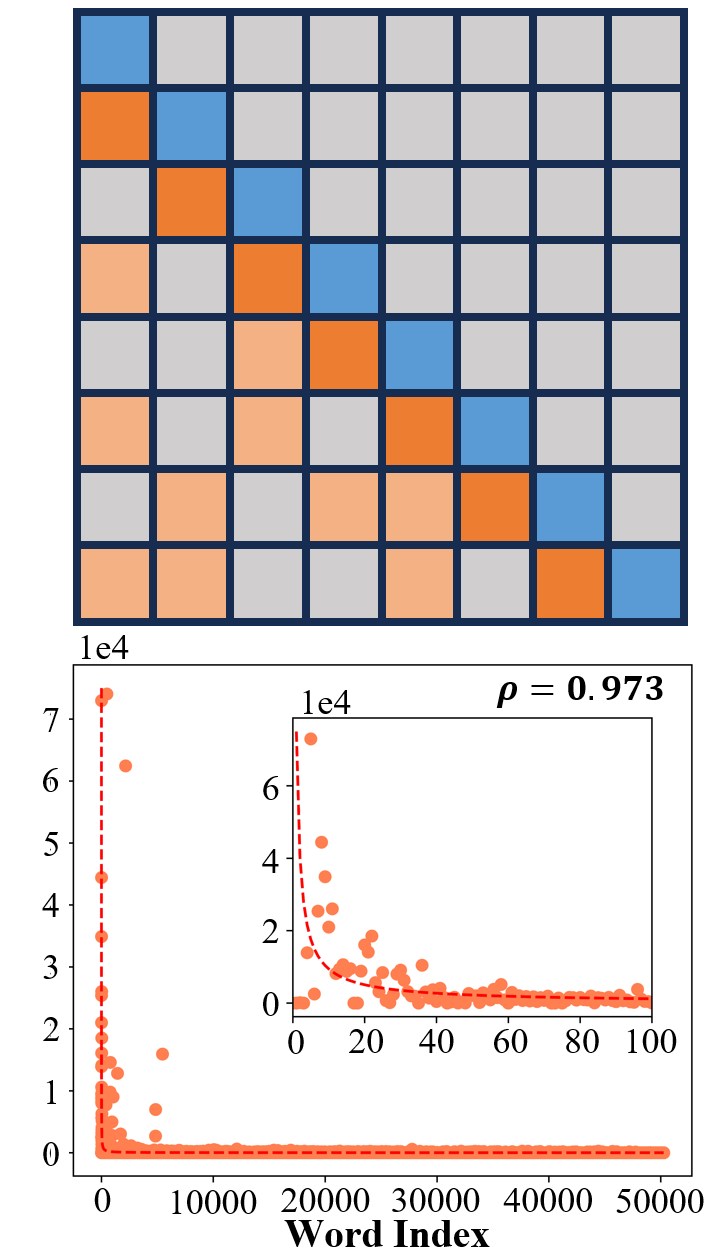}
         \label{fig:attn-1-swa}
     \end{subfigure}
     \vspace{-5mm}
        \caption{Comparisons of our proposed Sparse Window Attention (SWA) and existing methods. 
        On the top are illustrative sparse patterns for attention weight matrices generated by each method, where the x-axis the positions in the input sequence that are being attended to, and the y-axis represents the positions in the output sequence.
        The same notation is used in Figure~\ref{fig:attn-1}.
        Grey blocks mean the values are masked with zeros, due to the autoregressive LLM inference.
        On the bottom are the corresponding average attention score distributions in the Wiki-Text-2 dataset vocabulary for the OPT-6.7B model. 
        $\rho$ is the Spearman correlation score between sparse attention and dense attention (higher is better).
        }
        \label{fig:attn-2}
\end{figure*}
\begin{figure}[!t]
\begin{center}
  \includegraphics[width=\linewidth]{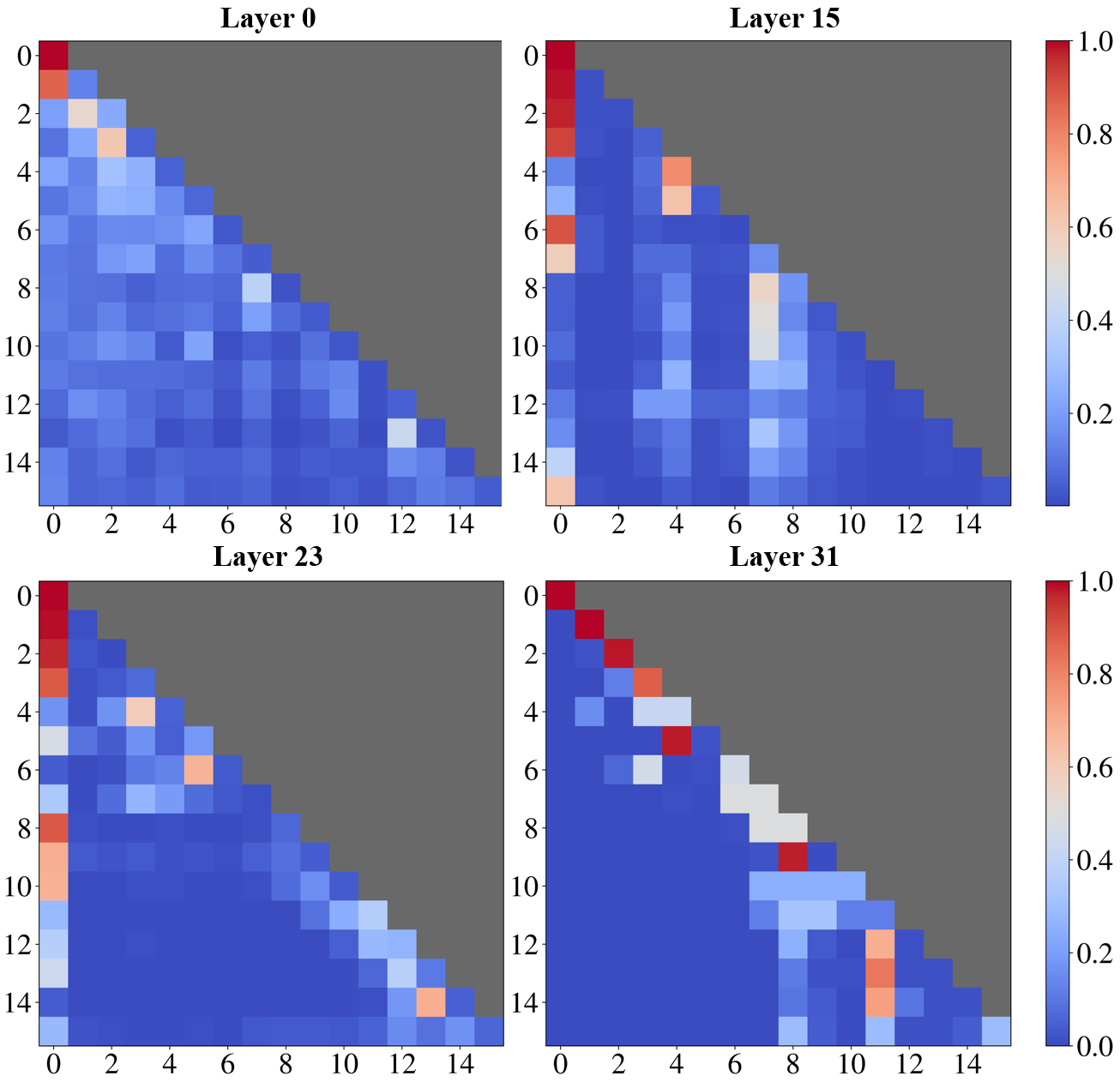}
\end{center}
\vspace{-2mm}
  \caption{Average attention weight maps for dense attention in OPT-6.7B on the Wiki-Text-2 dataset~\cite{wiki}.
  The sequence length is 16.
  Grey blocks mean the values are masked with zeros, due to the autoregressive LLM inference.
  }
\label{fig:attn-1}
\end{figure}
\begin{figure}[!t]
\begin{center}
  \includegraphics[width=\linewidth]{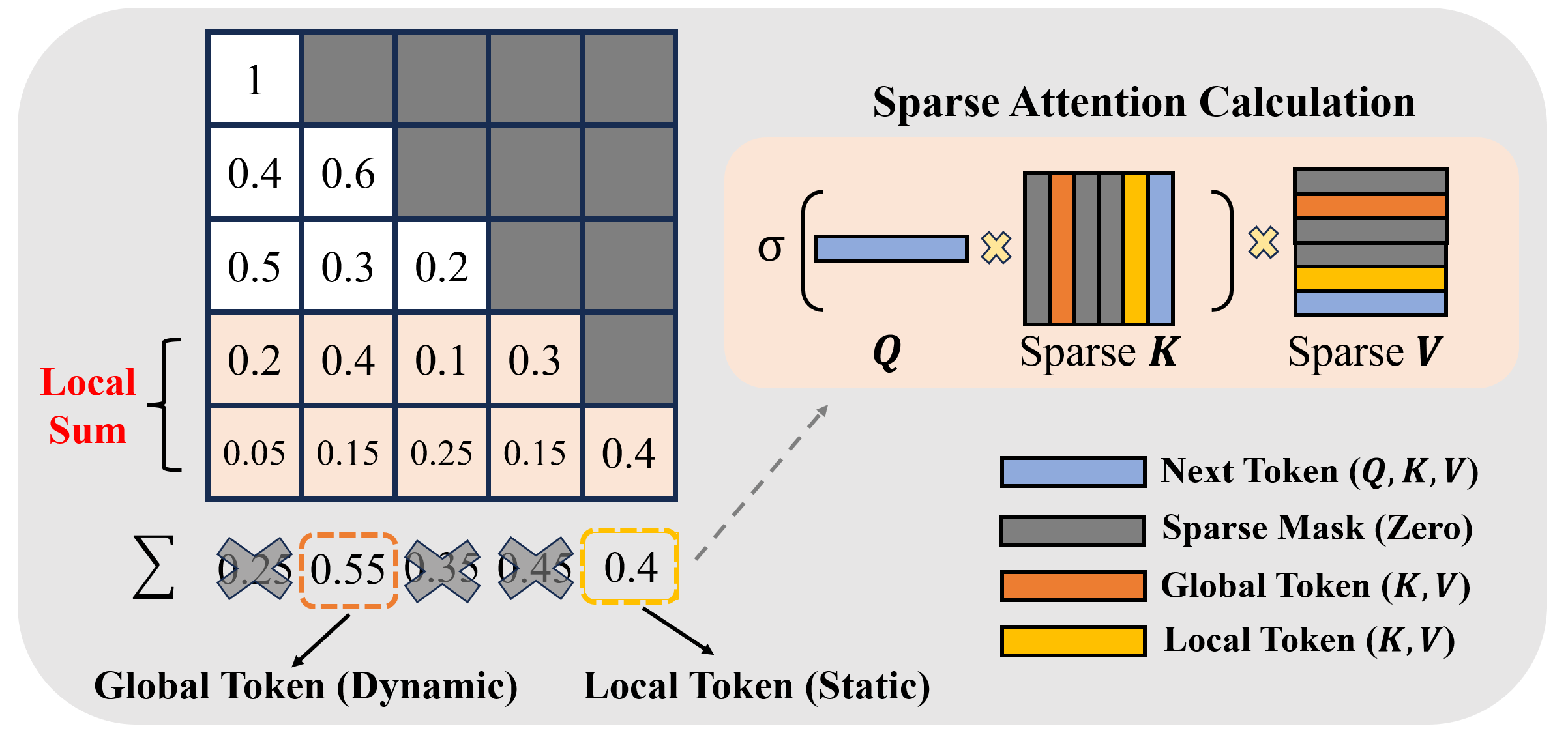}
\end{center}
\vspace{-2mm}
  \caption{
  An illustrative example of our proposed Sparse Window Attention (SWA) algorithm using 40\% caching ratio.
  The left matrix denotes the attention weight map.
  We calculate the local attention sum using solely the two most recent tokens.
  The locally static token is kept regardless of the value of the local attention sum.
  The globally dynamic token is selected as the one with the highest local attention sum.
  }
\label{fig:swa}
\end{figure}
\subsection{Objective}
To make the most of the high sparsity in attention weights, we propose to co-design LLMs in resource-constrained systems from both the algorithm and system sides.
Three technical questions need to be answered.

\minisection{Identifying Important Tokens.} 
In the context of LLMs, individual tokens have varying importance. 
During the inference process, the attention weights for each token vary from step to step.
The nondeterministic nature of language makes it extremely hard to predict which tokens are important.
Hence we need a low-cost mechanism to distinguish important tokens without hurting accuracy significantly for LLM inference.

\minisection{Caching KV Tensors.}
When KV tensors become too large for GPU memory, we have to store partial KV tensors in CPU memory for future reuse.
Theoretically, we could use Belady's Algorithm as the caching policy~\cite{belady}, which evicts the tokens that will not be used for the longest period in the future.
However, this oracle algorithm assumes future knowledge and imposes a huge amount of resources, making it impractical in LLM inference.
Therefore there is a need to develop a low-cost caching policy to allocate sparse KV tensors and ensure a relatively low miss rate.

\minisection{Caching vs. Recomputation.}
As the sequence length grows, the benefit of KV caching diminishes at a certain threshold since the time for accessing CPU memory might outweigh that for recomputing partial KV tensors. 
Moreover, this sequence length threshold varies across batch sizes and model configurations.
Thus, we must design a dynamic scheduling strategy that balances KV caching and recomputation at the token level.

\section{ALISA Algorithm Design}\label{sec:algorithm_design}
\subsection{Attention Analysis}
In the LLM era, existing works mainly aim to create sparsity in attention weights during LLM inference~\cite{longformer,sparseformer}.
Longformer~\cite{longformer} adopts a local attention mechanism, which applies a fixed-size sliding window on the KV tensors corresponding to the most recent tokens.
The resultant attention weight pattern is shown on the top of Figure~\ref{fig:attn-2}~(b).
SparseTransformer applies a strided mask on the tokens and creates strided attention~\cite{sparseformer}, as shown on the top of Figure~\ref{fig:attn-2}~(c).

To understand why the previous attention methods fail upon long sequences, we visualize the dense attention weight maps during LLM inference in Figure~\ref{fig:attn-1}.
We observe that attention weights with larger values do not exhibit a specific pattern.
Only using the most recent tokens cannot accurately represent the distribution of the entire attention weights, since the tokens with large attention weights (therefore more important) are often far from the current token.
A similar problem exists in strided attention, and the stride mask might not always capture large attention weights.
Therefore, the attention weight maps of the local attention and the strided attention can not capture a large portion of attention weights.
Subsequently, the corresponding attention score distributions significantly drift away from what is expected in dense attention.
At the bottom of Figure~\ref{fig:attn-2}~(a)-(c), we see that dense attention scores follow a near power-law distribution, which is consistent with previous findings~\cite{mongoose,linformer}. However, the attention score distributions generated by local and strided attention show close to zero correlation to that of dense attention, thus resulting in drastically lower accuracy.

\subsection{Sparse Window Attention (SWA)}

To maintain model accuracy, we propose a novel Sparse Window Attention (SWA) method, which produces both \textit{locally static} and \textit{globally dynamic} sparse patterns. 
We generate static patterns at local tokens by keeping the most recent tokens to preserve language sequential semantics and generate dynamic patterns to capture the dynamically changing semantic importance of prior tokens.
The importance of the prior tokens for future token generation is determined by the sum of the local attention weights.
Figure~\ref{fig:swa} draws an example of our SWA algorithm, with the algorithm details formulated as in Algorithm~\ref{alg:swa}.

\begin{algorithm}[!t]
\begin{algorithmic}[1]
\Require Previous attention weight $\textit{AW}$, query $Q$, keys and values $K, V$, caching ratio $r$, sequence length $n$, hidden dimension $d$. 
Note that this work evenly splits final tokens into $k$ globally dynamic and $k$ local static tokens.
The local attention sum is also reduced along the head dimension (not shown for conciseness).
\Ensure Attention score $\textit{Attn}$
\State $k = \nint{\frac{nr}{2}}$
\State $S = \sum {\textit{AW}}[n-k:n-1]$ \Comment{Local attention sum}
\State ${I}^{l} = [n-k, ..., n-1]$ \Comment{Locally static tokens}
\State ${I}^{g} = \text{argmax}_{k} {S}$ \Comment{Globally dynamic tokens}
\State $I = [{I}^{l}, {I}^{g}]$  \Comment{Sparse tokens}
\State ${K}_s, {V}_s = {K}[I,:], {V}[I,:]$ \Comment{Sparse KV tensors}
\State $\textit{AW} = \sigma(\frac{QK_{s}^{T}}{\sqrt{d}})$ \Comment{Attention weight}
\State $\textit{Attn} = \textit{AW} \cdot {V}_s$ \Comment{Softmax \& attention score} \\
\Return $\textit{Attn}$
\end{algorithmic}
\caption{\ourmethod's Sparse Window Attention}
\label{alg:swa}
\end{algorithm}
Our method is based on the hypothesis that multiple preceding steps can provide better hints on which tokens are more important than a single step.
The resultant sparse patterns are shown on the top of Figure~\ref{fig:attn-2}~(d).
Note that there do exist prior works that generate sparse patterns based on the entire attention weights~\cite{spatten,h2o}.
However, the entire attention weights will quadratically increase memory footprint with sequence length, thus not scalable upon long sequences.
We plot the distribution of the resultant attention scores at the bottom of Figure~\ref{fig:attn-2} (d).
Unlike previous fixed sparse patterns, our SWA produces a nearly identical power-law distribution as the dense attention and obtains a Spearman correlation score close to 1.
This similarity validates the efficacy of our SWA algorithm.
The details of SWA are formulated in Algorithm~\ref{alg:swa}.
Two key differences exist between dense attention and SWA.
First, the algorithm entails a caching ratio to determine how many tokens to keep at each step for KV sparsity and apply the sparse masks at the token level.
While irregular sparsity could exist across tokens, each token is still a dense tensor.
Second, we use gather operations to pack sparse KV tensors into a dense one and perform dense matrix operations.
Therefore, despite the multi-step attention calculation in SWA, both the computation and memory access for SWA remain regular, if we target a proper granularity.

\begin{figure*}[!ht]
\begin{center}
  \includegraphics[width=\linewidth]{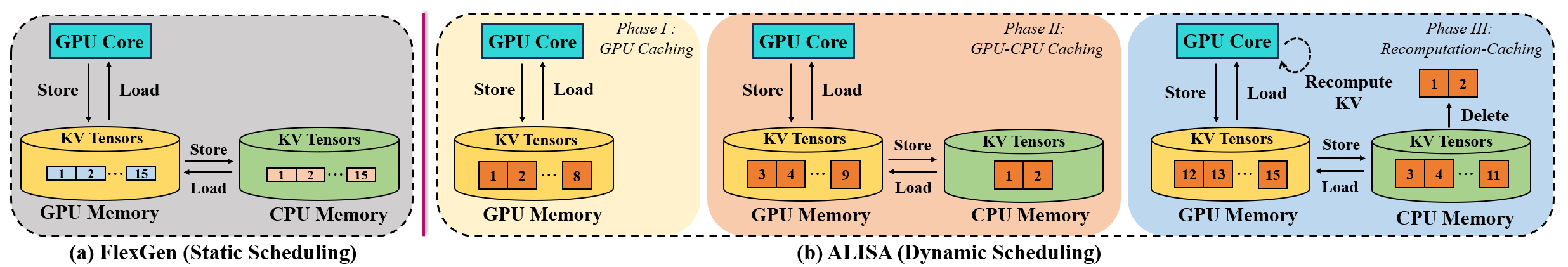}
\end{center}
\vspace{-3mm}
  \caption{(a) FlexGen's static scheduling.
  This scheduling splits KV tensors along the head dimension and remains static across different sequence lengths.
  (b) \ourmethod's dynamic scheduling. 
  In phase I, the entire KV tensors are small enough to fit in GPU memory, and no CPU memory access exists.
  In phase II, when GPU capacity is not enough for all KV tensors, the CPU is also used for caching preceding KV tensors.
  In phase III, recomputing partial KV tensors is faster than retrieving them from CPU memory. 
  These KV tensors are deleted from CPU memory and recomputed in GPU.
  The phase change is triggered by the sequence length, and the autoregressive inference of different tokens can be in different phases.
  }
\label{fig:ds}
\end{figure*}

\section{\ourmethod System Design}\label{sec:system_design} 
Since SWA introduces sparse KV tensors dynamically, designing a system for LLM inference that handles such sparsity effectively is essential.
We propose dynamic scheduling to ensure SWA lives up to its potential at the system level.
Then, we leverage KV compression to further improve the system-level performance.
Our proposed \ourmethod is a synergetic symbiosis of SWA, dynamic scheduling, and KV compression. 
Specifically, SWA identifies important KV tensors and generates sparse patterns. 
Then the dynamic scheduling utilizes important tokens and user-specified caching ratio to balance sparsity-aware caching and recomputation at the token level during LLM inference.
The KV compression further reduces the overall memory footprint of KV tensors by quantizing them into INT8 format.

\subsection{Dynamic Scheduling}
\minisection{Three-Phase Scheduling.}
\begin{algorithm}[!t]
\begin{algorithmic}[1]
\State \textbf{Initialization:} GPU core \textit{GC}, GPU memory \textit{GM}, CPU memory \textit{CM}, sequence length $n$, phase switch step $\{p_1, p_2\}$, offload ratio $\alpha$ and recompute ratio $\beta$ for KV tensors.
\For{all $j < n$} 
\\\textcolor{gray}{\# Load}
\If{$ j \geq p_{1}$} \Comment{Phase II \& III}
    \State \textit{CM}$\rightarrow$\textit{GM}.load($\ast$)
\EndIf
\State \textit{GM}$\rightarrow$\textit{GC}.load($\ast$) \Comment{Phase I\&II\&III}
\\\textcolor{gray}{\# Compute}
\If{$j \geq p_{2}$} \Comment{Phase III}
    \State Recompute($\ast$)
\EndIf
\State Update($K_{j}, V_{j}$) \Comment{Attention computation}
\\\textcolor{gray}{\# Store}
\State \textit{GC}$\rightarrow$\textit{GM}.store($K_{j}, V_{j}$) \Comment{Phase I\&II\&III}
\If{$j \geq p_{1}$} \Comment{Phase II}
    \If{$j \geq p_{2}$} \Comment{Phase III}
        \State \textit{CM}.delete($K^{\beta}_{j}, V^{\beta}_{j})$)
    \EndIf
    \State \textit{GM}$\rightarrow$\textit{CM}.store($K^{\alpha}_{j}, V^{\alpha}_{j}$)
\EndIf
\EndFor
\end{algorithmic}
\caption{\ourmethod's Dynamic Scheduling}
\label{alg:sys}
\end{algorithm}
Since the size of KV tensors gradually increases with longer sequences, it is evident that the engaged memory will increase over time.
Due to the high cost of CPU memory I/O accesses, one shall balance the memory access and computation at the token level to maximize the performance.
Our scheduling is described as follows.
\begin{itemize}
    \item Phase I: \textit{GPU Caching}. 
    All KV tensors can fit in GPU memory and are stored in the GPU.
    \item Phase II: \textit{GPU-CPU Caching}. 
    The total size of all KV tensors exceeds the capacity of GPU memory, and the KV tensors are split at the token level on both GPU and CPU memory and accessed upon need.
    \item Phase III: \textit{Recomputation-Caching}.
    After a certain sequence length, partial KV tensors are deleted from the CPU and recomputed in GPU if needed instead of being accessed from CPU memory.
\end{itemize}
We illustrate our scheduling with an illustrative example in Figure~\ref{fig:ds}~(b) and a formulation in Algorithm~\ref{alg:sys}.
Each inference pass contains load, compute, and store parts.
Load from GPU memory to GPU core is mandatory for all phases.
In Phase II and III, load from CPU to GPU happens before load from GPU memory to GPU core.
The new KV tensors will be computed and then stored in GPU memory.
In Phase II and III, certain KV tensors in GPU memory will be stored (offloaded) to CPU memory.
Since the global sparse patterns vary from step to step, we choose to keep the KV tensors for the locally static tokens in the GPU and store the preceding ones in the CPU.
Though there exist caching policies such as Belady's Algorithm~\cite{belady}, such oracle methods could be too computationally expensive to be impractical for efficient LLM inference.
Our heuristic-based caching policy can effectively reduce the potential CPU memory access with small enough compute overheads (compared to the memory bottleneck).
In Phase III, we delete the oldest KV tensors in the CPU and recompute these KV tensors in the GPU core when needed.

In contrast, prior works usually pre-defined static scheduling for KV tensors throughout the LLM inference~\cite{flexgen,h2o,vLLM}, as shown in Figure~\ref{fig:ds}~(a).
They fail to leverage the opportunity from dynamic memory capacity changes upon longer sequences, leading to sub-optimal performance.

\minisection{Sparsity-Aware Caching.}
The subsequent question is how to determine the phase switch step and offload and recompute ratio of KV tensors.
\begin{table}[!t]
    \centering
    \caption{Notations.}
    \begin{tabular}{c||l}
    \toprule
     $h, l, b$ & hidden dimension, layer count, batch size \\ \midrule
     $s, n$ &  input length, output length \\ \midrule
     $r, B$ & KV caching ratio, CPU-GPU bandwidth \\ \midrule
     $\alpha, \beta, p_1, p_2$ & offload/recompute ratio, phase switch step \\ \midrule
     $T^c, T^r$ & Time for compute and recompute \\ \midrule
     $T^m$ & Time for KV caching (CPU-GPU)  \\ \bottomrule
    \end{tabular}
    \label{tab:not}
\end{table}
We formulate this question as an optimization problem to minimize the total execution time.
We list the relevant notations in Table~\ref{tab:not}.
With FP16 format, the size of KV tensors for each token is $4 \cdot b \cdot l \cdot h$ bytes.
At a sequence length $j$, we denote the number of tokens moved from GPU to CPU as $\theta^{c}_{j}(\alpha) = \alpha(j+s)$ and the number of tokens moved from CPU to GPU as $\theta^{g}_{j}$.
The execution time of caching at step $j$ can be estimated as:
\begin{align}
    T^{m}_{j}(\alpha) &= \frac{4 \cdot b\cdot l \cdot h \cdot (\theta^{c}_{j} + \theta^{g}_{j})}{B} \\
    \quad & p_1 \leq j < n, \quad 0 \leq \theta^{g}_{j} \leq \nint{(s+j)r}
\end{align}

The optimization of the total execution time is formulated as:
\begin{align} 
    \min_{\{\alpha, \beta, p_1, p_2\}}  \quad & \sum_{j=1}^{p_2} T^{c}_{j} + \sum_{j=p_{1}}^{n} T^{m}_{j}(\alpha) + 
    \sum_{j=p_{2}}^{n} T^{r}_{j}(\beta) \\
    \mathrm{s.t.} \quad & 0 \leq p_{1} < p_{2} \leq n, 0 < \alpha < 1, 0 <\beta < 1
\end{align}
We solve this problem by dividing it into two sub-problems, including a data transfer problem and a computation problem.
The data transfer problem (the second term) is solved using hardware and software constraints, including memory capacity, bandwidth, KV tensor size, etc.
Conversely, the computation problem (the first and third terms) is solved via profiling.
We profile the execution time for compute and recompute with different configurations and create a mapping between input configurations and their execution time.
Then, we apply a greedy search method to solve the optimization problem for the best performance.
This process is done offline, introducing no overhead during LLM inference.
\begin{figure*}[!ht]
\begin{center}
  \includegraphics[width=\linewidth]{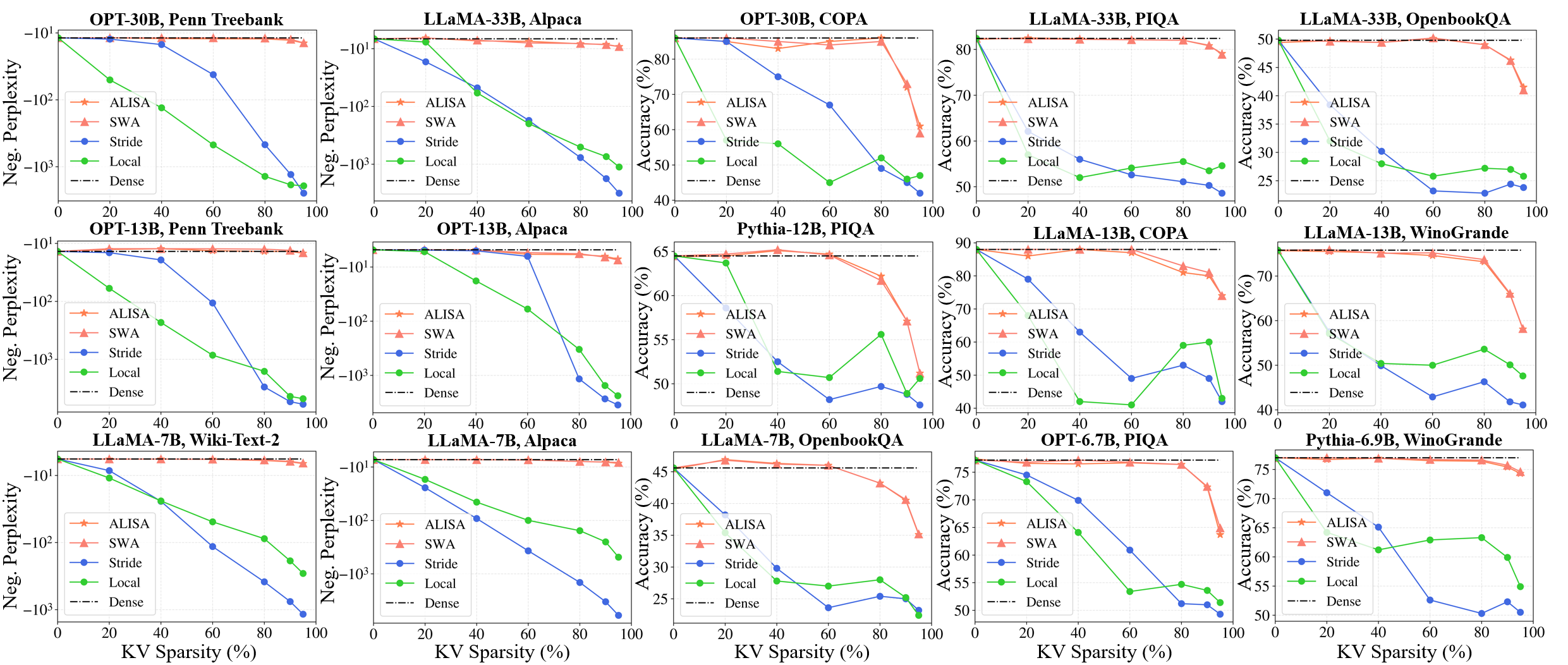}
\end{center}
\vspace{-3mm}
  \caption{Accuracy of \ourmethod (SWA + Compression), SWA, dense attention, local attention~\cite{longformer}, and strided attention~\cite{sparseformer}. 
  Along the y-axis, we arrange the measurements to be higher is better. 
  The input length is set to 2048 for all the datasets to match the maximal context length of each LLM.
  Negative perplexity and accuracy are utilized to measure the language modeling and question-answering tasks, respectively. 
  }
\label{fig:acc}
\end{figure*}
\subsection{KV Compression}
Previous works have utilized quantization to accelerate attention computation by compressing model weights~\cite{GPTQ,AWQ}.
In this work, we leverage quantization for a different purpose, i.e., compressing KV tensors to reduce memory access. 
We adopt a fine-grained channel-wise quantization for KV tensors for better inference robustness~\cite{robustquantization}. 
More specifically, we use the following formula to quantize KV tensors to $b$-bit integers in memory and de-quantize them to their original format (FP16 in this work) for computation:
\begin{align}
    x_{\text{quant}} = \text{round}(\frac{1}{\lambda}x + z), \quad x= \lambda(x_{\text{quant}} - z) 
\end{align}
where the scaling factor $\lambda=\frac{max - min}{2^b - 1}$, and zero point $z= \text{round}(\frac{-2^b}{max - min})$. 
Previous work finds that for OPT model can be compressed up to INT4 while maintaining accuracy~\cite{4bit}.
In this work, we choose to quantize KV tensors to INT8 to ensure our KV compression can be generalized to more LLMs.

\section{Evaluation}\label{sec:evaluation}

\subsection{Experimental Setup}
\minisection{Models and datasets.}
We use three open-sourced families of LLM models: OPT with 6.7B, 13B, and 30B parameters~\cite{opt}, LLaMA with 7B, 13B, and 33B parameters~\cite{llama2}, and Pythia with 6.7B and 12B parameters~\cite{pythia}. 
For algorithm-related evaluations, we use the lm-evaluation-harness library~\cite{lm-eval} and perform two popular language-related tasks on seven different datasets, namely language modeling for Wiki-Text-2~\cite{wiki}, Penn Treebank~\cite{ptb} and Alpaca~\cite{alpaca}, and 4-shot question-answering inference for PIQA~\cite{piqa}, COPA~\cite{copa}, OpenBookQA~\cite{openbookqa}, and Winogrande~\cite{winogrande}. 
We report task-specific metrics, e.g., perplexity for language modeling and accuracy for question-answering, across different model types and scales.
We set the input length as 2048, matching the maximal context length for LLMs, to showcase the algorithmic performance (perplexity and accuracy in this work) when operating at full context.

For the system evaluation, we sample and tokenize inputs from the Alpaca dataset. 
We use an input sequence length of 128 and an output sequence length of 512 to test our system under varying batch size configurations, ranging from 4 to 64.
We aim to evaluate the LLM system performance at all possible model scales, unless the configuration is not available (e.g., Pythia-30B does not exist).
The evaluation metric for performance is token throughput, defined as the end-to-end execution time (both prefilling and decoding stages) divided by the total number of generated tokens.
\begin{figure*}[!t]
\begin{center}
  \includegraphics[width=\linewidth]{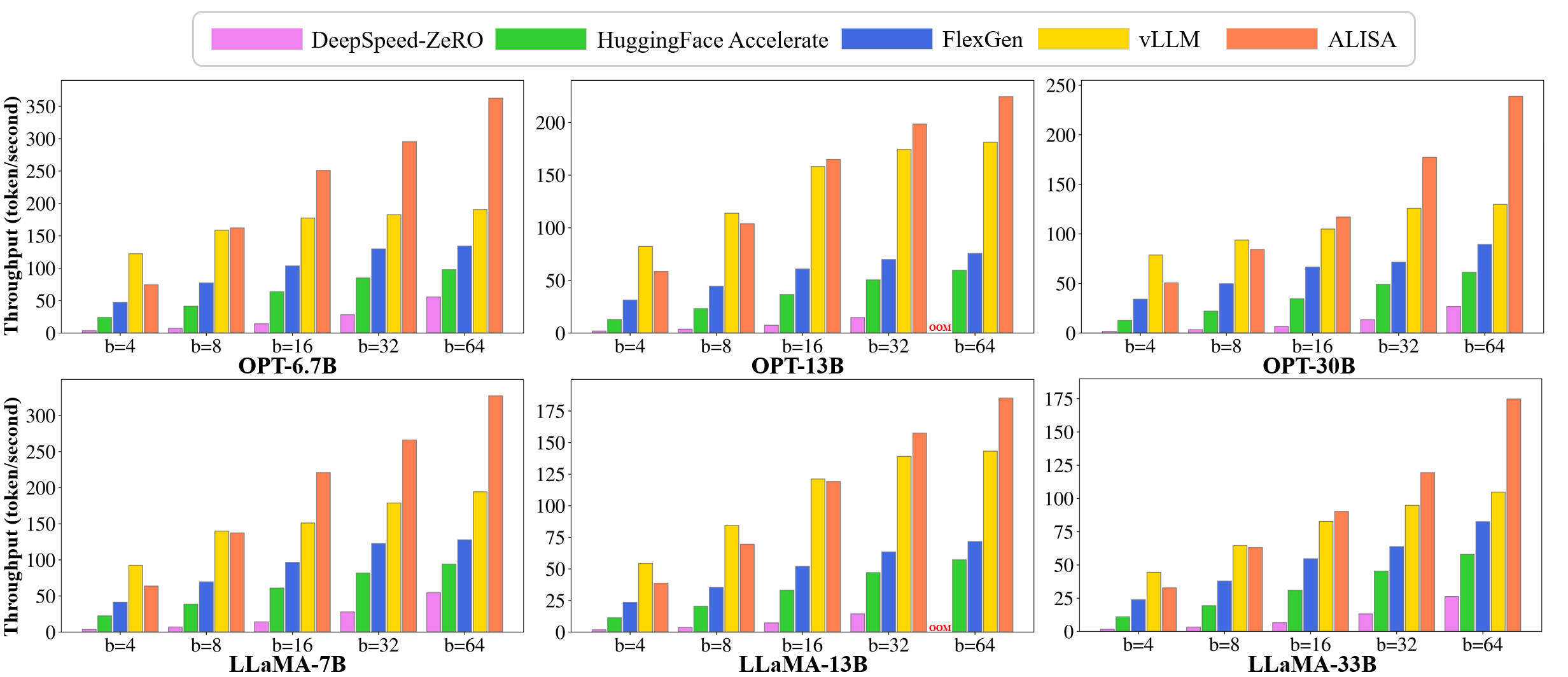}
\end{center}
\vspace{-3mm}
  \caption{Throughput of \ourmethod with 80\% KV Sparisty and baselines, including DeepSpeed-ZeRO~\cite{deepspeed}, HuggingFace Accelerate~\cite{huggingface}, FlexGen~\cite{flexgen}, and vLLM~\cite{vLLM} on the Alpaca dataset~\cite{alpaca}. 
  Along the y-axis, higher measurements are better. 
  OOM denotes out-of-memory error.
  We use an input length of 128 and an output length of 512. 
  }
\label{fig:sys}
\end{figure*}

\minisection{Baselines.}
To validate the accuracy, we use dense attention, local attention~\cite{longformer}, and strided attention~\cite{sparseformer} as our baselines. 
For system experiments, we use DeepSpeed-ZeRO~\cite{deepspeed}, HuggingFace Accelerate~\cite{huggingface}, and  FlexGen~\cite{flexgen}, and vLLM~\cite{vLLM}as baselines. 
DeepSpeed-ZeRO is a deep learning optimization software developed to improve the computation and memory efficiency of training and inference for large models. 
For LLM inference, DeepSpeed-ZeRO performs offloading weights instead of intermediate KV tensors. 
HuggingFace Accelerate is another open-sourced library that focuses on promoting easy and reproducible transformer-based research. 
It supports offloading the whole KV tensors to the CPU memory during LLM inference. 
FlexGen is a very recent LLM-specific work that focuses on optimizing LLM inference in single GPU-CPU systems.
It defines a static scheduling allocation strategy by solving an offline linear programming problem to minimize the total execution time given the memory constraints.
vLLM is a dedicated online LLM inference serving system for multi-tenant user requests~\cite{vLLM}.
It manages the KV tensors at the block level (fixed group of tokens). 
Each block is stored in non-contiguous paged memory and is swapped between CPU and GPU memory.

\minisection{Implementation.} 
We conduct our experiments in single GPU-CPU systems.
For 7B/13B level models, we use NVIDIA Tesla V100 with 16/32~GB HBM; for 30B level models, NVIDIA H100 with 80~GB HBM.
The CPU is 2.60~GHz Intel Xeon with 128~GB DRAM, and the bandwidth between GPU and CPU is 20~GB/s. 
\ourmethod is implemented on top of FlexGen~\cite{flexgen} and HuggingFace Transformers~\cite{transformers}. 
FlexGen allows users to offload model weights, KV tensors, and activations simultaneously.
Since we focus on optimizing KV caching, we keep both the model weights and activations always in GPU memory. 
In terms of memory allocation, we manage the memory space at the token level and schedule KV tensors in a layerwise manner. 
We use the FP16 format for all variables, except the KV compression.
\begin{figure}[!t]
\begin{center}
  \includegraphics[width=\linewidth]{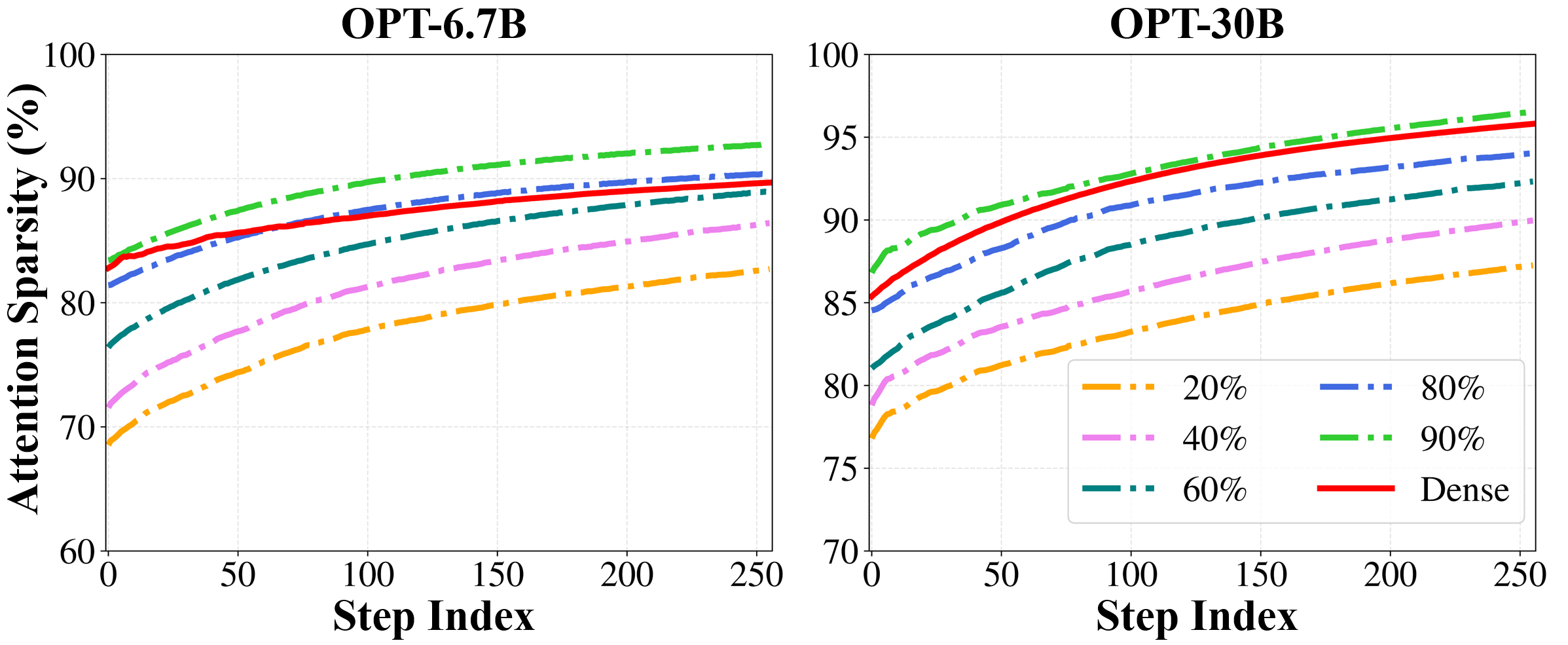}
\end{center}
\vspace{-4mm}
  \caption{
  Attention weight sparsity (averaged across all layers) upon different KV sparsity.
  We consider elements as zeros if they fall below 1\% of the row-wise maximum value. 
  }
\label{fig:sps-comparison}
\end{figure}
\begin{figure}[!t]
\begin{center}
  \includegraphics[width=\linewidth]{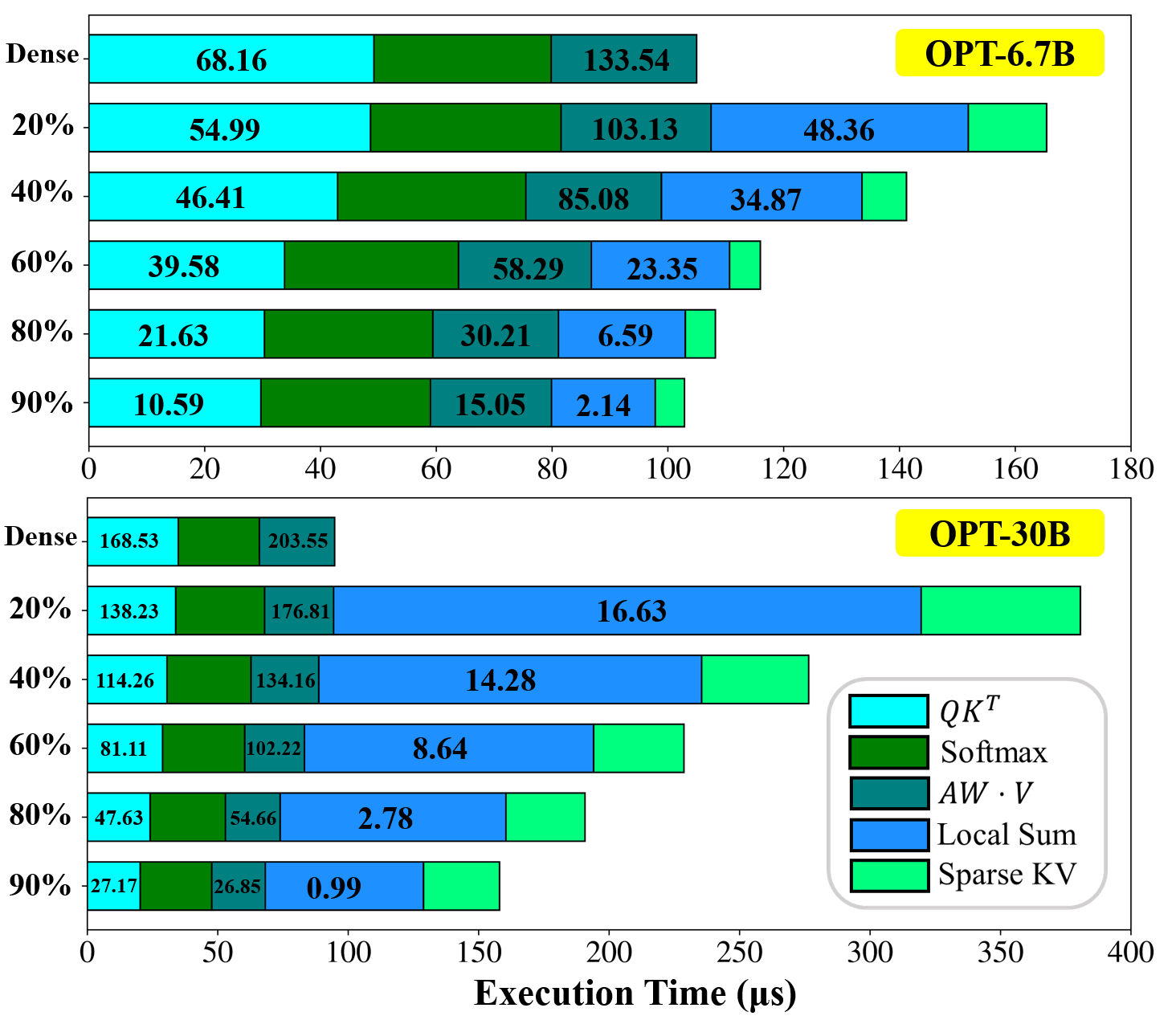}
\end{center}
\vspace{-3mm}
  \caption{
  Execution time of a single attention module. 
  Numbers within the bar indicate the corresponding floating point operations per second (FLOPS), either MAC or ADD.
  We use a batch size of 64 and a sequence length of 128.
  }
\label{fig:attn-breakdown}
\end{figure}
\begin{figure*}[!t]
\begin{center}
  \includegraphics[width=\linewidth]{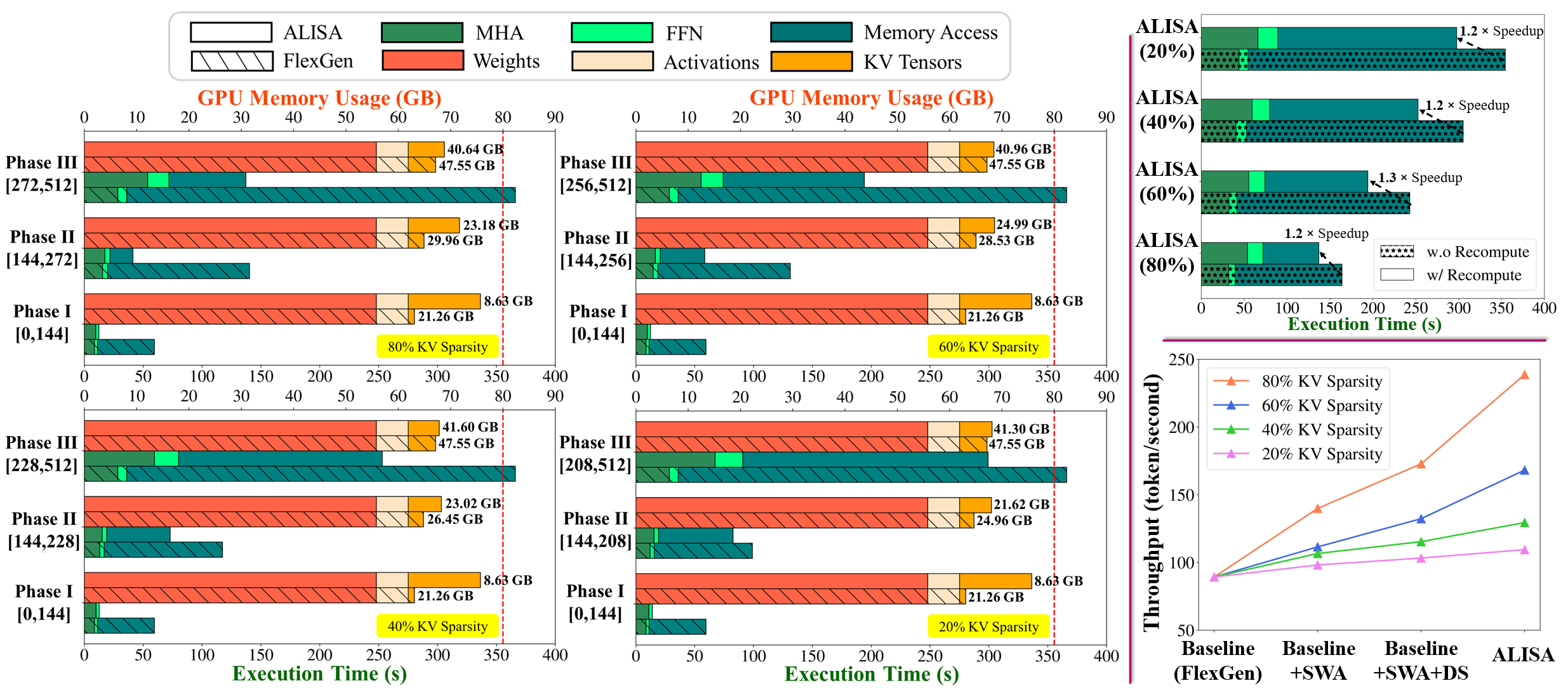}
\end{center}
\vspace{-4mm}
  \caption{LLM inference. 
  All experiments are conducted using OPT-30B with a batch size of 64, input length of 128, and output length of 512 on one NVIDIA H100 GPU. 
  All bars correspond to the sequence length at the end of the phase.
  (a) Left four: execution time and memory usage of FlexGen and \ourmethod by our proposed scheduling phase for different KV sparsity. 
  Numbers on the right-hand side of the memory bar indicate CPU memory usage and the red-dot line denotes the GPU memory capacity.
  (b) Top right: impact of recomputation on the execution time for different KV sparsity at the full sequence length.
  (c) Bottom right: ablation study on the impact of techniques for different KV sparsity. 
  DS denotes dynamical scheduling.
  }
\label{fig:breakdown}
\end{figure*}
\subsection{Accuracy}
We evaluate the accuracy for different KV sparsity, with results given in Figure~\ref{fig:acc}.
We observe that \textit{\ourmethod has consistent and significant improvements over local and strided attention methods} across different model types, model sizes, and datasets, demonstrating the effectiveness of \ourmethod.
We summarize three key insights here.
First, \ourmethod is a much more robust sparse attention method for LLMs.
For example, for LLaMA-33B on the top right, the accuracy of local and strided attention collapses instantly when sparse attention is adopted, i.e., the largest accuracy drop occurs from 0\% to 20\% KV sparsity, while \ourmethod almost maintains an identical accuracy to that of dense attention up to 80\% KV sparsity.
Second, \ourmethod is more robust when LLMs become larger.
With \ourmethod, fewer accuracy collapses occur at 80\% KV sparsity when LLM sizes increase from the 7B level to 13B/30B level, regardless of the model families.
However, no such trends exist in local and strided attention.
Third, KV compression is scalable with almost no accuracy impact for LLMs.
In all settings, we see that the accuracy of \ourmethod almost perfectly tracks that of SWA.
In certain settings, ALISA can even outperform dense attention, since well-structured sparsity can often act as regularization to improve accuracy on unseen datasets~\cite{spsreg1,spsreg2}.
Though minor discrepancies exist in OPT-30B and LLaMA-13B on the COPA dataset, we conclude that KV compression is practically scalable with KV sparsity and model size.
\subsection{Performance}
\minisection{End-to-end Throughput.}
We further evaluate the end-to-end system performance (i.e., throughput) of \ourmethod. 
We choose 80\% as the evaluated KV sparsity, as it is the maximum value for ALISA to retain good algorithmic performance, i.e., perplexity and accuracy, less than 5\% drop for most tasks shown in Figure~\ref{fig:acc}.
Specifically, the drop for the Alpaca dataset is around 3\%.
Figure~\ref{fig:sys} shows the performance of OPT and LLaMA models on the Alpaca dataset.
\textit{Overall, \ourmethod offers the highest attainable throughput for LLM inference in resource-constrained systems.} 
There are three key observations.
First, \ourmethod achieves consistent speedup over all baselines, showing $1.4\sim3.0\times$ higher throughput over FlexGen.
Prior works like DeepSpeed-ZeRO are not fully optimized for LLM inference by introducing out-of-memory errors upon large batch sizes, since it does not offload KV tensors.
Second, \ourmethod is more scalable than previous works.
As the batch size grows, the speedup of \ourmethod over FlexGen and other methods increases. 
Third, \ourmethod can sustain up to $1.9\times$ improvement over vLLM under larger batch sizes.
This is due to two reasons: 
1) \ourmethod co-designs the sparsity patterns and KV caching to reduce the memory footprint, while vLLM only optimizes the memory management of KV tensors; 
2) the dynamic scheduling strategy in \ourmethod features recomputation to further alleviate the memory bottleneck upon large KV tensors.
Note that under small batch sizes, vLLM outperforms as it is optimized for online serving with fine-grained memory management.

\minisection{Attainable Sparsity.}
We further show why \ourmethod can achieve this speedup.
Figure~\ref{fig:sps-comparison} shows the achieved sparsity after SWA.
Two key observations exist.
First, for both LLMs, allowing more sparse KV tensors will increase the sparsity in attention weights.
Second, for larger LLMs, we need a higher KV sparsity to close the gap between the attainable sparsity in our SWA and dense attention.
However, the accuracy with higher KV sparsity will likely drop according to Figure~\ref{fig:acc}.
Overall, the insight is that \ourmethod can reasonably take advantage of the opportunities in sparse attention to accelerate LLM inference.

\minisection{Breakdown of Attention Module.}
To better understand the impact of SWA, we profile the execution time of key operations in Algorithm~\ref{alg:swa}, with results given in Figure~\ref{fig:attn-breakdown}.
There are two key observations here.
First, SWA introduces an execution overhead, which varies across different KV sparsity.
Higher KV sparsity in SWA always reduces the execution time.
The main sources of reduction are the process of $QK^T$, local attention sum, and sparse KV tensors (i.e., using sparse token indices to generate dense KV tensors).
Larger LLMs incur higher overheads, especially in local attention sum and sparse KV tensors.
The reason is that larger LLMs have larger model dimensions.
For example, the hidden dimension and head number increase from $[4096,32]$ in OPT-6.7B to $[7168, 56]$ in OPT-30B.
This larger overhead also validates our argument that prior works that generate sparse attention based on the entire attention weight are not scalable~\cite{spatten,h2o}.
Second, there exists under-utilization in the $QK^T$ computation for SWA. 
The corresponding execution time does not decrease proportionally as KV sparsity increases, leading to a significant FLOPS drop.
The main reason is that a smaller dense tensor gathered from sparse KV tensors can not fully utilize massive parallel GPU cores.
The execution time for the local sum scales with KV sparsity.
Higher sparsity is induced by a lower caching ratio, which in turn reduces the number of additions in the local sum. 
However, the local sum could spend more time than $QK^T$ computation, due to its low data use, i.e., vector vs. matrix operation.

\minisection{Breakdown of LLM Inference.}
Figure~\ref{fig:breakdown} shows the details of the full LLM inference.
There are three key observations in Figure~\ref{fig:breakdown}~(a).
First, \ourmethod is always faster than FlexGen for all KV sparsity and all phases.
With higher KV sparsity, the speedup of \ourmethod over FlexGen is more significant.
Both the time spent on computation and memory access is less when KV sparsity goes higher since fewer KV tensors are involved per step.
However, the main contributor to reducing the execution time is that fewer KV tensors need to be moved between CPU and GPU.
As we have both statically local and dynamically global sparse patterns, we prefer allocating local tokens in GPU to reduce CPU memory access, since global tokens are less predictable.
Second, \ourmethod always makes better use of the GPU memory than FlexGen in all cases.
The total memory requirement of all KV tensors (the total GPU and CPU memory usage) increases with the sequence length.
The GPU memory usage is not directly related to KV sparsity, as our dynamic scheduling optimizes the execution time instead of GPU memory usage.
Third, the size of KV tensors indeed impacts when a phase starts.
Different KV sparsity leads to varying tensor sizes and triggers Phase III at different sequence lengths, and higher KV sparsity enters Phase III later.
\ourmethod in Phase III has a smaller size of KV tensors than FlexGen due to deleting partial KV tensors.
Overall, \ourmethod manages KV caching at the token level and balances the caching and recomputation in a more fine-grained manner than FlexGen.
Figure~\ref{fig:breakdown}~(b) studies the impact of recomputation in Phase III.
We observe that recomputation can reduce the total execution time by $1.2\sim1.3\times$.
Though recomputation induces additional computation overhead, it results in a more substantial reduction in execution time due to decreased memory accesses.
Figure~\ref{fig:breakdown}~(c) shows the ablation study.
Across different KV sparsity, we observe that different techniques almost contribute equally, and the gain of each technique increases proportionally with the KV sparsity.

\section{Conclusion}\label{sec:conclusion}
In this work, we present an algorithm-system co-design solution, \ourmethod, to accelerate LLM inference in resource-constrained systems. 
On the algorithm level, \ourmethod adopts a Sparse Window Attention (SWA) algorithm to create a mixture of globally dynamic and locally static sparse patterns and reduces the memory footprint with negligible accuracy degradation. 
On the system level, \ourmethod leverages a three-phase scheduler to dynamically allocate KV tensors and achieves optimal throughput by balancing caching and recomputation. 
Experiments show that, in single GPU-CPU systems, \ourmethod achieves up to $3\times$ and $1.9\times$ throughput improvement over FlexGen and vLLM, respectively.


\bibliographystyle{IEEEtranS}
\bibliography{refs}

\end{document}